\documentclass[letterpaper,journal]{IEEEtran}
\usepackage{amsmath,amsfonts,amssymb}
\usepackage{algorithmic}
\usepackage{algorithm}
\usepackage{array}
\usepackage{multirow}
\usepackage[caption=false,font=normalsize,labelfont=sf,textfont=sf]{subfig}
\usepackage{textcomp}
\usepackage{stfloats}
\usepackage{url}
\usepackage{verbatim}
\usepackage{graphicx}
\usepackage{cite}
\usepackage{hyperref}
\hypersetup{colorlinks=true,linkcolor=blue,anchorcolor=blue,citecolor=blue}

\usepackage[capitalize]{cleveref}
\crefname{section}{Sec.}{Secs.}
\Crefname{section}{Section}{Sections}
\Crefname{table}{Table}{Tables}
\crefname{table}{Tab.}{Tabs.}

\begin{document}

\bstctlcite{BSTcontrol}

\title{Behavior-Skill: A Fine-Grained Benchmark for Evaluating Vision-Language-Action Policies in Long-Horizon Tasks}


\author{
Chunyun Ma$^{1,2}$,
Lun Luo$^{2}$,
Xingjian Luo$^{3,2}$,
Xiexing Feng$^{2}$,
Hang Zhang$^{2}$,
Wei Liu$^{2}$,
Feng Qiao$^{2}$,
Yaonan Wang$^{4}$,
Huimin Lu$^{1}$,
and Xieyuanli Chen$^{1}$%
\thanks{
$^{1}$College of Intelligence Science and Technology and National Key Laboratory of Equipment State Sensing and Smart Support, National University of Defense Technology, Changsha, China.
$^{2}$XPeng Inc., Guangzhou, China.
$^{3}$The Chinese University of Hong Kong, Hong Kong, China.
$^{4}$Hunan University, Changsha, China.
Corresponding author: Xieyuanli Chen (chenxieyuanli@hotmail.com).
}}




\maketitle

\begin{abstract}
Reliable execution of long-horizon mobile manipulation tasks remains challenging because overall task success depends on the successful completion of multiple constituent skills. Existing benchmarks, however, still rely primarily on full-task rollouts and aggregate task-level metrics, making intermediate failures difficult to observe and analyze. We present Behavior-Skill, a benchmark that reformulates the learning and evaluation of long-horizon tasks around executable constituent skills. It contains 235,492 skill instances from 10,000 demonstrations across 50 household tasks and 34 semantic skill categories. Each instance pairs a skill instruction with an aligned observation-action segment, and is further associated with a restorable intermediate state and a skill success condition to enable independent evaluation under valid preconditions. We further introduce trajectory-level and skill-level metrics to characterize policy capability beyond aggregate task success. Extensive experiments across representative VLA policies including $\pi_{0.5}$ and GR00T on the complete 50-task benchmark  show that failures are highly non-uniform across skills, with contact-rich manipulation skills forming persistent bottlenecks. These results demonstrate that Behavior-Skill complements full-task evaluation by exposing intermediate capability profiles for analyzing and improving long-horizon VLA policies. Behavior-Skill is publicly available at
\url{https://github.com/nubot-nudt/Behavior-Skill}. 
\end{abstract}


\section{Introduction}

\IEEEPARstart{I}{n} Long-horizon mobile manipulation tasks, robots are expected to perform increasingly complex activities such as household assistance, warehouse automation, and industrial assembly. As they must continuously integrate perception, language understanding, planning, and manipulation over extended interaction horizons, developing policies that can reliably accomplish such long-horizon tasks has become a critical challenge~\cite{sapkota2025vla_survey}.

\begin{figure}[!t]
\centering
\includegraphics[width=\columnwidth]{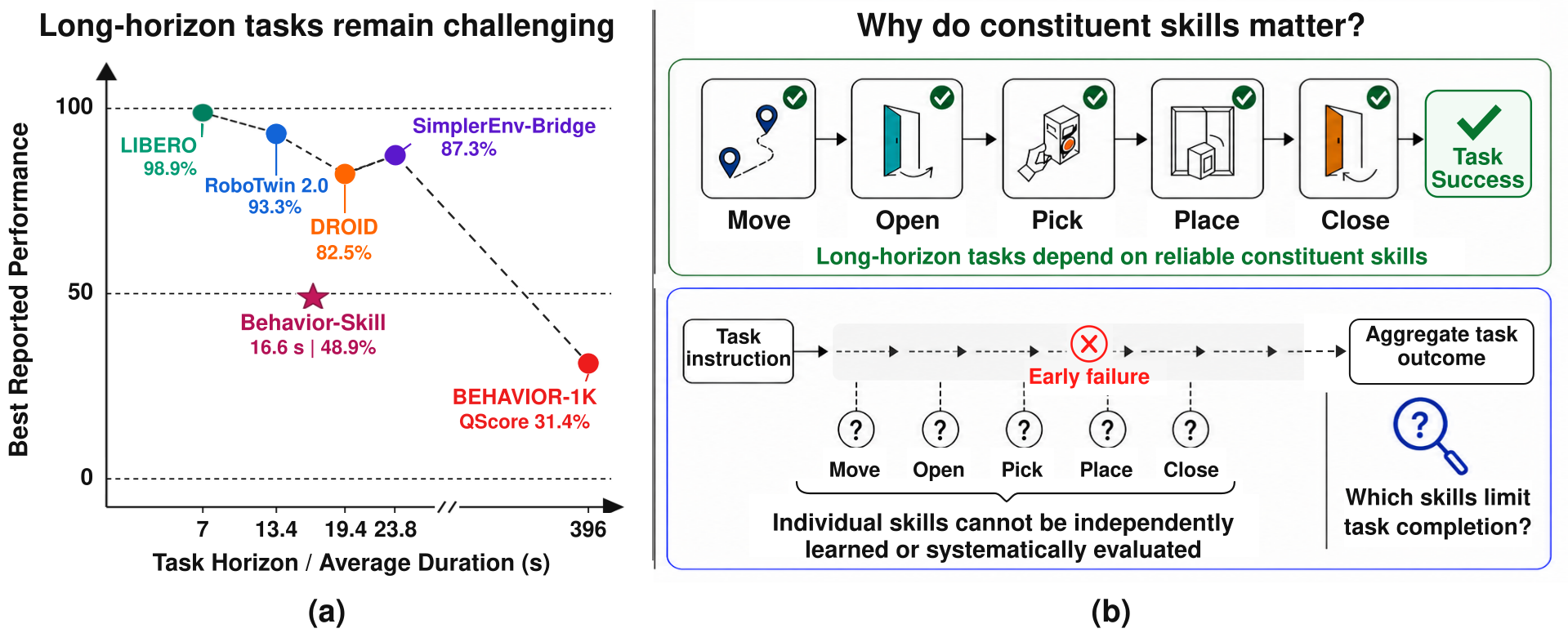}
\caption{Challenges of long-horizon mobile manipulation tasks.
\textbf{(a)} Performance on representative embodied benchmarks with increasing task horizon, measured by average task duration. The Behavior-Skill result denotes the overall skill success rate in our evaluation, while the other results use their respective benchmark metrics.
\textbf{(b)} Long-horizon tasks require reliable execution of multiple constituent skills.}
\label{fig: head_fig}
\vspace{-0.6cm}
\end{figure}

Recent Vision-Language-Action (VLA) models have made remarkable progress toward this goal by learning unified policies that directly map visual observations and language instructions to robot actions. Large-scale robot pretraining and open-source foundation models, including RT-2~\cite{zitkovich2023rt2}, OpenVLA~\cite{kim2025openvla}, $\pi_0$-series~\cite{black2025pi0,black2025pi05}, GR00T\cite{bjorck2025grootn1} and Gemini Robotics~\cite{gemini2025robotics}, have substantially improved policy generalization across objects, environments, and robots. On several widely adopted embodied benchmarks~\cite{liu2023libero,li2024evaluating,khazatsky2024droid,chen2025robotwin}, these models have achieved increasingly competitive performance as shown in ~\cref{fig: head_fig}a. However, performance on challenging long-horizon benchmarks such as BEHAVIOR-1K~\cite{li2023behavior1k} remains considerably lower~\cite{galaxea2026g05,bai2025openpicomet,larchenko2025behaviorchallenge}, indicating that executing reliable complex multi-stage tasks is still far from solved.
  
The capability to accomplish long-horizon tasks fundamentally depends on the reliable execution of multiple constituent skills. As shown in ~\cref{fig: head_fig}b,  a task can only be completed when its intermediate skills are successfully executed sequentially, and the failure of any skill will  interrupt task progress. Consequently, improving long-horizon VLA policies requires understanding and strengthening constituent skill execution. However, existing embodied datasets and benchmarks~\cite{li2023behavior1k,mees2022calvin,liu2023libero,zhang2025vlabench} are primarily organized at the task level. They provide only task-level language annotation, and evaluation is performed through continuous task rollouts using final task success or overall progress metrics. As a result, constituent skills can hardly be independently learned or systematically evaluated, and it remains unclear which skills actually limit long-horizon task completion. These limitations become increasingly pronounced as task horizons grow, where early failures may prevent later skills from being executed and aggregate task metrics provide only limited information for understanding policy capability.

To address these challenges, we present Behavior-Skill, a fine-grained benchmark that establishes constituent skills as the basic unit for studying long-horizon VLA policies built on the Behavior-1K~\cite{li2023behavior1k} dataset. Behavior-Skill provides a skill dataset together with an independent evaluation framework, enabling systematic learning and evaluation of constituent skills under a unified experimental setting. Extensive experiments on representative VLA models including $\pi_{0.5}$~\cite{bai2025openpicomet} and GR00T~\cite{bjorck2025grootn1} reveal that failures in long-horizon tasks are concentrated in a few semantic skill categories rather than being uniformly distributed across complete task trajectories. These findings suggest that reliable long-horizon task execution is largely constrained by a limited set of bottleneck skills, highlighting the importance of skill-centric data and evaluation for future long-horizon VLA research.
 
In summary, the main contributions of this work are summarized as follows: 
\begin{itemize}
\item{We  present Behavior-Skill, a fine-grained benchmark for long-horizon VLA policies with 235,492 skill instances across 34 semantic skill categories, establishing constituent skills as the fundamental unit for learning and evaluation. 
}
\item{We establish an independent skill evaluation framework together with capability-oriented metrics, enabling systematic measurement of constituent skills under valid preconditions while preserving the original task context.
} 
\item{We conduct extensive experiments across representative VLA backbones, revealing that failures in long-horizon mobile manipulation tasks are highly non-uniform, with contact-rich skills as bottlenecks.}

\end{itemize}

\section{RELATED WORK}
Achieving reliable long-horizon task execution has attracted increasing attention in embodied intelligence. Existing research mainly advances this problem from two complementary directions: developing more capable Vision-Language-Action policies and constructing increasingly challenging benchmarks.

\subsection{Vision-Language-Action Models for Long-Horizon Tasks} 
Recent progress in long-horizon mobile manipulation tasks has been largely driven by Vision-Language-Action (VLA) models trained on large-scale robot demonstrations. Representative models, including RT-1~\cite{brohan2023rt1}, Octo~\cite{octo2024}, OpenVLA~\cite{kim2025openvla}, $\pi_0$ series~\cite{black2025pi0,black2025pi05} and GR00T~\cite{bjorck2025grootn1}, substantially improve generalization across tasks, environments, and robot embodiments. More recent studies further explore adaptive reasoning~\cite{lin2025onetwovla}, action tokenization~\cite{pertsch2025fast}, parameter-efficient adaptation~\cite{kim2025openvlaoft}, spatial representations~\cite{qu2025spatialvla} and agentic planning~\cite{yang2025agenticrobot} to improve long-horizon policy execution. However, as shown in \cref{fig: head_fig}a, recent VLA policies typically achieve over 80\% performance on representative short-horizon benchmarks, whereas the best reported performance on BEHAVIOR-1K is only 31.4\%~\cite{galaxea2026g05}, highlighting the difficulty of reliable long-horizon task execution.

To better model long-horizon mobile manipulation tasks, many studies introduce fine-grained representations between task-level instructions and low-level robot actions. Representative approaches employ subgoals~\cite{ahn2023saycan}, executable programs~\cite{liang2023codeaspolicies}, fine-grained action and process representations~\cite{shridhar2020alfred,wang2025worldmodeling,wu2025robocoin} to support planning, reasoning, and language-guided policy learning. More recent VLA methods further incorporate intermediate step-wise instructions and instruction-oriented training to  strengthen reasoning-action alignment during long-horizon execution~\cite{shi2025hirobot,yang2025instructvla}. These studies have shown the potential of fine-grained representations for policy learning. However, existing work primarily treats skills as supervision signals or planning abstractions, while a unified framework that supports both learning and independent evaluation of constituent skills has not been systematically established.

\subsection{Evaluation of Long-Horizon Tasks}
Benchmark development has played an important role in advancing long-horizon mobile manipulation tasks. CALVIN~\cite{mees2022calvin}, LIBERO~\cite{liu2023libero}, ARNOLD~\cite{gong2023arnold}, VLABench~\cite{zhang2025vlabench}, RoboCerebra~\cite{han2025robocerebra} and BEHAVIOR-1K~\cite{li2023behavior1k} provide increasingly challenging multi-stage tasks for evaluating policy generalization and sequential task execution. Complementary frameworks such as SIMPLER~\cite{li2024evaluating}, Colosseum~\cite{pumacay2024colosseum}, and REALM~\cite{sedlacek2025realm} further investigate robustness, sim-to-real consistency, and cross-environment generalization. Complementary to benchmark development, recent studies have explored fine-grained evaluation through structured skill stages~\cite{wang2025roboeval}, execution-quality assessment~\cite{liu2026trustworthy}, subgoal-level evaluation~\cite{elmallah2025score}, and policy failure diagnosis~\cite{sagar2026robomd}. 

As summarized in Table~\ref{tab:benchmark-compare}, representative long-horizon benchmarks predominantly evaluate policies through complete task rollouts using aggregate task-level metrics.  Although recent studies provide finer-grained analysis of execution, they largely retain the original task-level evaluation protocol without independently executing constituent skills under valid execution preconditions. Consequently, constituent-skill capabilities cannot be systematically measured or directly compared across policies.  

\begin{table}[htbp]
\caption{Comparison of Representative Benchmarks for Robot VLA and Their Evaluation Protocols\label{tab:benchmark-compare}}
\centering
\begin{tabular*}{\columnwidth}{@{\extracolsep{\fill}}lcccccc}
\hline
Benchmark      & \shortstack[c]{Multi-\\stage} & \shortstack[c]{Skill\\Ann.} & \shortstack[c]{State\\Reset} & \shortstack[c]{Skill\\Eval.} & \shortstack[c]{Local\\Goals} & \shortstack[c]{Skill\\Metric} \\ \hline
CALVIN~\cite{mees2022calvin}         & $\checkmark$             & $\circ$                & $\times$                    & $\circ$                          & $\checkmark$        & $\times$          \\
LIBERO~\cite{liu2023libero}         & $\circ$                  & $\times$                & $\times$                    & $\times$                           & $\times$        & $\times$          \\
ARNOLD~\cite{gong2023arnold}         & $\circ$                  & $\times$                & $\times$                    & $\times$                           & $\circ$        & $\times$          \\
VLABench~\cite{zhang2025vlabench}       & $\checkmark$             & $\circ$                & $\times$                    & $\times$                           & $\circ$             & $\times$          \\
BEHAVIOR-1K~\cite{li2023behavior1k}    & $\checkmark$             & $\circ$                 & $\times$                    & $\times$                           & $\circ$        & $\times$          \\
REALM~\cite{sedlacek2025realm}          & $\times$                  & $\circ$                & $\times$                    & $\circ$                            & $\circ$             & $\circ$             \\ \hline
Behavior-Skill    & $\checkmark$ & $\checkmark$ & $\checkmark$ & $\checkmark$ & $\checkmark$ & $\checkmark$ \\ \hline
\end{tabular*}
\vspace{6pt}
\begin{minipage}{\columnwidth}
\footnotesize
\textit{"Multi-stage" indicates support for complex multi-stage tasks; "Skill Ann." denotes internal skill annotations; "State Reset" denotes intermediate-state restoration; "Skill Eval." indicates independent skill evaluation; "Local Goals" denotes local symbolic goals; "Skill Metric" indicates skill-type metrics. $\checkmark$, $\circ$, and $\times$ denote full, partial, and no support, respectively.}
\end{minipage}
\end{table}

\section{Behavior-Skill Benchmark}
\begin{figure*}[!t]
\centering
\includegraphics[width=\textwidth]{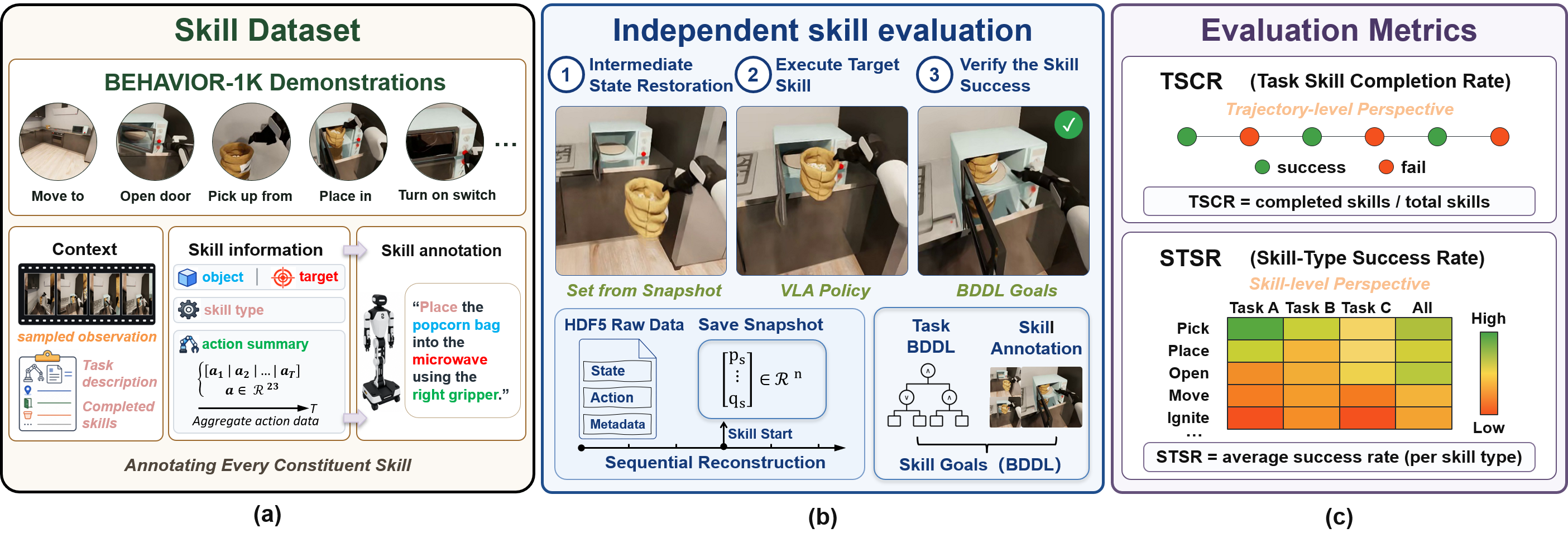}
\caption{\textbf{Overview of Behavior-Skill.} \text{(a)} Skill Dataset: we annotate each constituent skill from BEHAVIOR-1K demonstrations with multimodal context and action summaries. \textbf{(b)} Independent skill evaluation: each skill is executed from a restored intermediate state and verified against a skill-specific BDDL goal. \textbf{(c)} Evaluation metrics: TSCR summarizes skill completion at the trajectory level, while STSR captures performance across skill types.}

\label{fig:overview}
\vspace{-0.4cm}
\end{figure*}

Behavior-Skill builds upon BEHAVIOR-1K~\cite{li2023behavior1k}, establishing constituent skills as the fundamental unit for studying long-horizon mobile manipulation tasks. As shown in Fig.~\ref{fig:overview}, we first build a skill dataset by annotating every constituent skill. In addition, we provide an independent evaluation benchmark that restores valid intermediate states and evaluates constituent skills under satisfied preconditions, allowing constituent skills to be measured independently of preceding trajectory outcomes. Lastly, we propose capability-oriented metrics that summarize execution performance from complementary trajectory-level and semantic skill-type perspectives.

\subsection{Skill Dataset}
\label{subsec:skill_dataset}

As illustrated in \cref{fig:overview}a, we construct a skill dataset by annotating each skill of Behavior-1K. Each annotation describes the action, involved objects, spatial relation, execution context, and end-effector usage of the current skill. 

We first construct the textual description context required for each skill annotation from the original BEHAVIOR-1K skill records. These records specify the action label, involved object identifiers, manipulated object, skill type, and temporal interval. For example, a record may specify the action label \texttt{place in}, the object identifiers \texttt{popcorn\_bag\_73} and \texttt{microwave\_hjjxmi\_0}, and the manipulated object \texttt{popcorn\_bag\_73}. These structured fields describe the current skill but do not provide the global task background or temporal context. To recover the global task background and temporal dependencies between constituent skills, we additionally include the task description together with the previously completed skill descriptions. However, these textual fields cannot describe the visual interaction process or how the robot physically executes the skill.

To address this limitation, we augment the textual annotation context with synchronized visual observations and robot actions. The synchronized multi-view observations provide complementary information about the scene configuration and manipulation process. However, directly using the original videos is inefficient because each skill contains a different number of frames and three camera streams are recorded independently. We therefore design a duration-adaptive visual sampling strategy. For each skill interval, frames are sampled according to the skill duration, with at most 64 timestamps retained. The sampled head, left-wrist, and right-wrist views are resized and concatenated into a single multi-view observation, preserving both global scene context and local end-effector interactions. Visual observations may still leave arm and gripper usage ambiguous, especially under occlusion and coordinated bimanual manipulation. We therefore further incorporate the aligned robot actions. Each demonstration provides a frame-level 23-DoF action sequence containing base motion, two 7-DoF arm commands, and two gripper commands. These actions are recorded at the control frequency and contain dense low-level commands, making them unsuitable as direct annotation input. We therefore aggregate the actions within each skill interval into a compact motion description, e.g., \texttt{left arm moving, gripper closing; right arm moving, gripper open}. This motion summary provides explicit execution evidence for subsequent language annotation.

\begin{figure}[!htbp]
\centering
\includegraphics[width=\columnwidth]{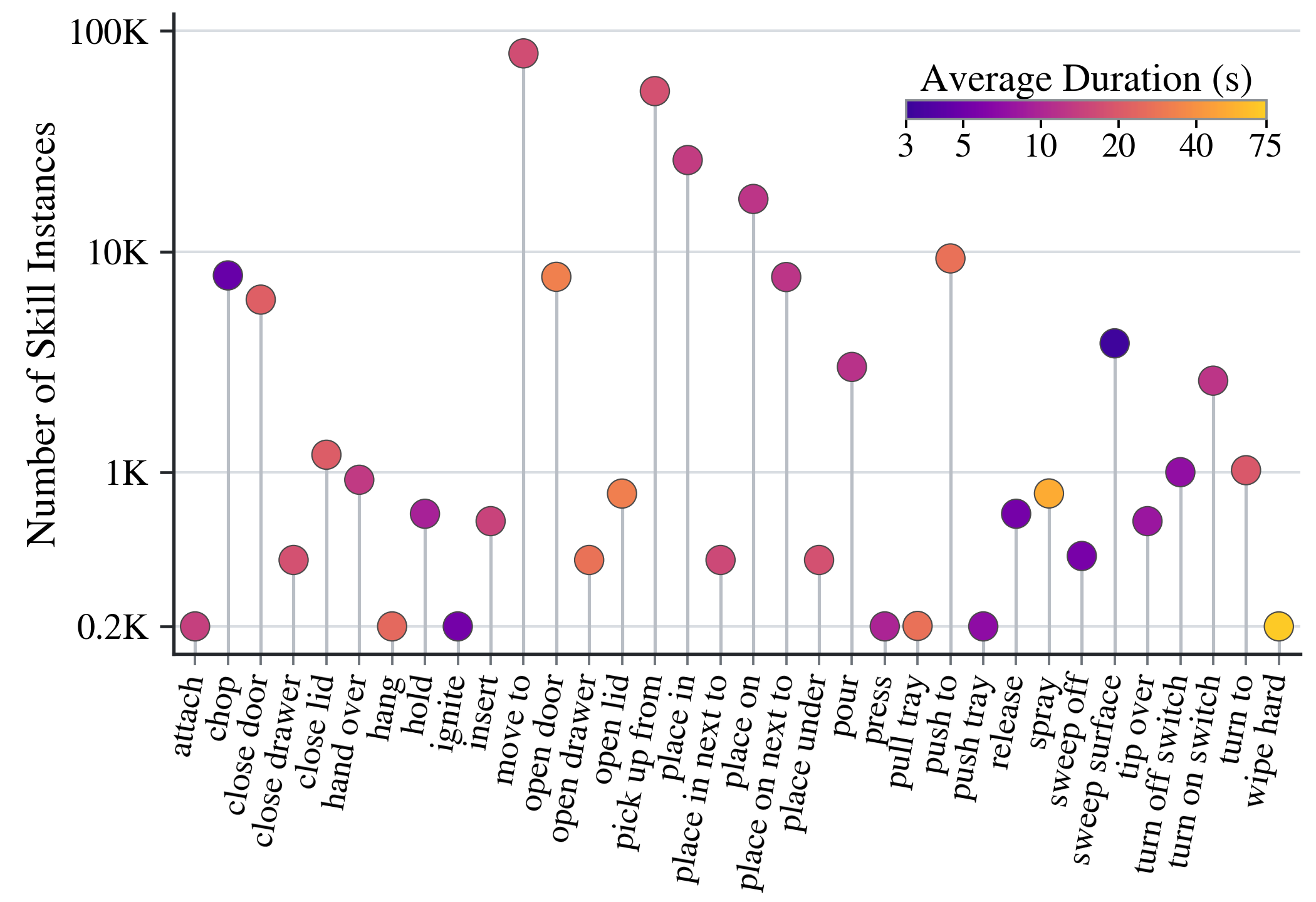}
\caption{Distribution of  skill instances in Behavior-Skill. The number of instances is shown for each semantic skill category, while marker color indicates the corresponding average execution duration. }
\label{fig:data_static}
\vspace{-0.4cm}
\end{figure}

Finally, we provide the constructed multimodal context to Qwen3-VL-235B-A22B-Instruct~\cite{bai2025qwen3vl} to generate annotation for the current constituent skill. For each task, we first manually inspect one representative demonstration to establish a reference annotation. The annotations of remaining demonstrations are then compared against this reference using ChatGPT-5 to identify inconsistencies in object identities, skill sequences, and annotation wording. Flagged cases are manually reviewed and corrected when necessary, followed by representative inspection of the final annotations. The resulting Behavior-Skill provides 235,492 skill instances with an average execution duration of 16.6 s. They are constructed from 10,000 demonstrations spanning all 50 long-horizon BEHAVIOR-1K tasks. \cref{fig:data_static} shows the distribution of the 34 semantic skill categories and their average execution durations, which range from 3.2 s to 70.9 s, illustrating the diversity of constituent skills represented in the dataset.

\subsection{Independent Skill Evaluation}
\label{subsec:skill_evaluation}
The objective of Behavior-Skill is not only to provide skill data for policy learning, but also to establish a unified protocol for independently evaluating constituent skills under valid execution conditions. The proposed framework complements complete-task rollouts by measuring whether a policy can correctly execute an individual constituent skill when its execution preconditions are satisfied. This allows intermediate capabilities to be directly measured without being obscured by failures accumulated during preceding stages.

As shown in \cref{fig:overview}b, independent evaluation additionally requires an executable definition of successful skill completion. Therefore, the skill instance introduced in the previous Section is further associated with evaluation-specific information to construct an executable evaluation unit,

\begin{equation}
E_{i,j}
=
\left(
S_{i,j},
L_{i,j},
G_{i,j},
H_{i,j}
\right)
\label{eq:evaluation_unit}
\end{equation}

\noindent where $S_{i,j}$ denotes the simulator state immediately before
executing the $j$-th skill of the $i$-th demonstration,
$L_{i,j}$ is the corresponding skill instruction,
$G_{i,j}$ is the skill success condition, and $H_{i,j}$ is
the maximum evaluation horizon. 

\textbf{Intermediate State Restoration.}
For each constituent skill, the simulator state $S_{i,j}$ in Eq.~\eqref{eq:evaluation_unit} is defined as an intermediate scene snapshot. It serves as the initialization state for independent skill evaluation while preserving the original execution context within the task.

These snapshots are constructed from the original OmniGibson HDF5~\cite{li2023behavior1k} demonstration recordings provided as the raw demonstrations of BEHAVIOR-1K. These recordings capture the complete simulation history collected during human teleoperation, including frame-wise serialized states, control actions, and environment transition events. However, the recorded frame states use an active-object filter, where active objects are those that are awake or whose object states were updated during the current simulation step. Sleeping objects without state updates may therefore be omitted, making an isolated frame generally insufficient to recover the complete intermediate scene configuration.  To address this limitation, a trajectory-based state reconstruction procedure is performed. Starting from the initial state of each demonstration, the recorded transition events are applied sequentially, the corresponding serialized state is loaded, and the recorded action is executed to advance the physics engine. This procedure continues until the annotated starting frame of the target skill, at which point the complete simulator state is serialized as an intermediate scene snapshot. Assisted-grasp constraints are also reconstructed before serialization to preserve object attachments established during the original demonstration.

The resulting intermediate scene snapshot stores all information required to faithfully reconstruct the execution state of the constituent skill. Specifically, it records the simulator version information, scene construction parameters, robot configuration, articulated-object states, object poses, interaction context, and assisted-grasp constraints.

\textbf{Skill Success Specification.}
The skill success condition $G_{i,j}$ in Eq.~\eqref{eq:evaluation_unit} is defined as a symbolic goal expressed in the Behavior Domain Definition Language (BDDL)  ~\cite{srivastava2022behavior,li2023behavior1k}. It serves as the success criterion for independent skill evaluation. BDDL represents goals as logical combinations of object-state predicates, where each predicate evaluates the current simulator state and determines whether a specified object relation or state is satisfied. To construct the corresponding skill success condition, we jointly analyze the original task BDDL definition together with the corresponding skill annotation to identify the manipulated objects, target objects, interaction type, and execution semantics. Based on this information, we manually define a symbolic goal that captures only the completion criterion of the current constituent skill while preserving the original object scope and logical representation.  

We construct each skill goal using either existing BDDL predicates or skill-specific extensions.  Most manipulation skills can be directly represented using existing BDDL predicates describing placement, articulated-object states, spatial relations, and object activation. However, some skills describe intermediate robot behaviors rather than terminal environment states. Navigation, grasping, and handover are representative examples, as their completion cannot be fully determined from the original task-level goals.  We therefore introduce a small set of skill-specific predicate extensions to represent behaviors that are not directly supported by the original BDDL predicates. These extensions capture robot-object proximity, grasp relations, handover relations, and other interaction conditions required by individual skills. Each predicate directly evaluates the corresponding geometric or interaction condition from the current simulator state and returns a Boolean result through the same interface as native BDDL predicates, allowing seamless integration with the original BDDL goal engine. When a skill requires several conditions to hold, the corresponding predicates are combined into a single goal through logical conjunction.

\textbf{Evaluation Procedure.}
Each evaluation unit $E_{i,j}$ in \cref{eq:evaluation_unit} is independently evaluated from its restored simulator state. First, the simulator is initialized by restoring the intermediate scene snapshot $S_{i,j}$ to recover the beginning state of the skill. Then the corresponding skill goal $G_{i,j}$ is assembled into a complete BDDL problem definition and injected as the active evaluation goal. The policy receives the current observations together with the corresponding skill instruction $L_{i,j}$ and executes the target skill. After each interaction step, the BDDL goal engine evaluates the current simulator state against the injected skill goal. The evaluation horizon $H_{i,j}$ is set to twice the recorded duration of the corresponding skill in the original demonstration, providing additional execution time for learned policies. The evaluation terminates immediately once all goal predicates are satisfied. Otherwise, execution continues until $H_{i,j}$ is reached, after which the skill is recorded as unsuccessful.

\subsection{Evaluation Metrics}

Independent skill evaluation produces a binary execution outcome for every constituent skill. While these binary outcomes directly indicate whether skills succeed or fail, they do not provide a quantitative summary of policy performance across complete demonstrations or semantic skill categories. To summarize these outcomes from complementary perspectives, Behavior-Skill reports two evaluation metrics as shown in \cref{fig:overview}c. Task Skill Completion Rate (TSCR) measures skill completion within each demonstration, whereas Skill-Type Success Rate (STSR) characterizes execution performance across semantic skill categories.

\textbf{Task Skill Completion Rate (TSCR).} 
To quantify how completely a policy executes the constituent skills required by a demonstration, we define Task Skill Completion Rate (TSCR). Let $\tau_i$ denote the execution trajectory of the $i$-th demonstration, 
containing $M_i$ constituent skills, and let
$y_{i,j}\in\{0,1\}$ indicate whether its $j$-th skill is
successfully executed.
The Task Skill Completion Rate of $\tau_i$ is defined as

\begin{equation}
\operatorname{TSCR}(\tau_i)
=
\frac{1}{M_i}
\sum_{j=1}^{M_i}
y_{i,j}
\label{eq:trajectory_tscr}
\vspace{-0.2cm}
\end{equation}

The overall benchmark performance is reported as the average TSCR over all evaluated demonstrations,

\begin{equation}
\overline{\operatorname{TSCR}}
=
\frac{1}{N}
\sum_{i=1}^{N}
\operatorname{TSCR}(\tau_i)
\label{eq:benchmark_tscr}
\vspace{-0.2cm}
\end{equation}

\noindent where $N$ denotes the total number of evaluated demonstrations.

\vspace{0.5em}

\textbf{Skill-Type Success Rate (STSR).}
Although TSCR quantifies skill completion from the perspective of an individual task, it does not aggregate execution performance across skills with the same semantics. We therefore define Skill-Type Success Rate (STSR) to measure the execution success rate of each semantic skill category.

For a skill category $k$, the Skill-Type Success Rate is
\begin{equation}
\operatorname{STSR}_k
=
\frac{
\displaystyle
\sum_{i=1}^{N}
\sum_{j=1}^{M_i}
\mathbb{I}\!\left(c_{i,j}=k\right)y_{i,j}
}{
\displaystyle
\sum_{i=1}^{N}
\sum_{j=1}^{M_i}
\mathbb{I}\!\left(c_{i,j}=k\right)
}
\label{eq:stsr}
\end{equation}

\noindent where $c_{i,j}$ denotes the semantic category of the $j$-th
skill in the $i$-th demonstration, and
$\mathbf{I}[\cdot]$ is the indicator function.

Together, TSCR and STSR summarize policy performance from complementary trajectory-level and semantic skill-type perspectives.

\section{EXPERIMENTS}
\subsection{Experimental Setup}

\textbf{Evaluation settings.} We conduct the primary evaluation on the complete 50-task Behavior-Skill benchmark using two representative VLA policies, $\pi_{0.5}$~\cite{black2025pi05} and GR00T N1.7~\cite{bjorck2025grootn1}. For each task, we randomly reserve ten demonstrations as the evaluation set, while the remaining demonstrations are used for training. This results in 500 evaluation demonstrations in total, and all skills contained in the selected demonstrations are included in the evaluation.  For each policy, we train task and skill variants using the same demonstrations, model architecture, observation and action representations, optimization objective, and training schedule. The two variants differ only in the language condition (task prompt, skill prompt).

\textbf{Training details.} All $\pi_{0.5}$ variants are initialized from the official pretrained $\pi_{0.5}$ checkpoint and use an action horizon of 32. The models are trained for 50,000 steps with a cosine learning-rate schedule, a peak learning rate of $2.5\times10^{-5}$, and a per-device batch size of 192. GR00T N1.7 is initialized from the official pretrained checkpoint and fine-tuned for 150,000 steps using a per-device batch size of 256, a learning rate of $1\times10^{-4}$, and a warm-up ratio of 0.05. All other training settings are kept identical between the Task and Skill variants within each policy architecture.

\textbf{Evaluation protocol.} All experiments adopt the proposed independent skill evaluation framework, where each constituent skill is executed from its restored intermediate state under satisfied preconditions. We consider two experimental settings throughout the paper: Task, which denotes policies trained with task instructions and conditioned on the original task prompt during inference, and Skill, which denotes policies trained with skill instructions and conditioned on the corresponding skill prompt. Apart from the language condition, both settings share identical restored intermediate states, skill success conditions, and execution horizons.

\subsection{Overall Skill Capability}
\begin{figure*}[!t]
\centering
\includegraphics[width=\textwidth]{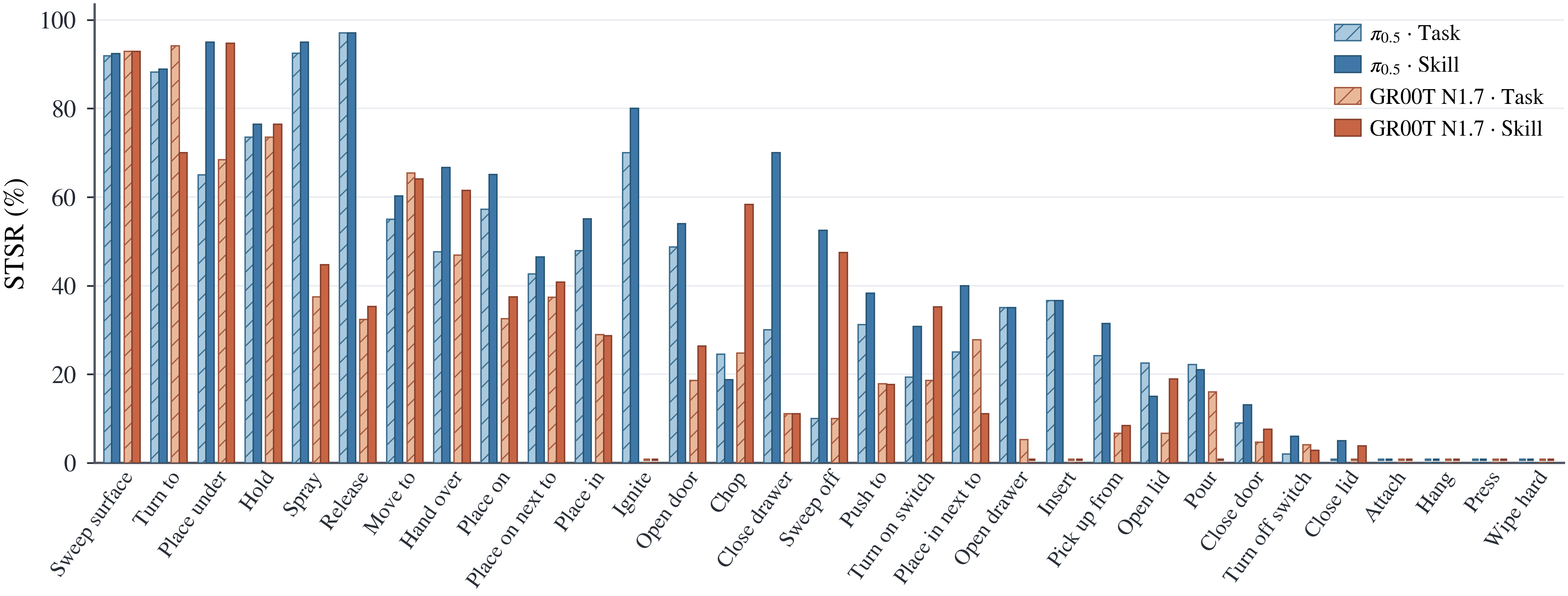}
\caption{Capability profiles (STSR, \%) of $\pi_{0.5}$ and GR00T N1.7 across semantic skill categories under Task and Skill settings on the complete 50-task benchmark.}
\label{fig:over_skill_performance}
\vspace{-0.4cm}
\end{figure*}

We evaluate constituent-skill capability on the complete 50-task Behavior-Skill benchmark using the proposed TSCR and STSR metrics. Table~\ref{tab:overall_tscr_50task} reports the overall TSCR, while Fig.~\ref{fig:over_skill_performance} presents the STSR of each semantic skill type. Under the Skill setting, $\pi_{0.5}$ and GR00T N1.7 achieve TSCRs of only 48.4\% and 42.5\%, respectively. Thus, even after skill training, the two evaluated policies successfully execute fewer than half of the constituent skills on average. These results indicate that skill execution remains a major limitation for long-horizon task execution.

\begin{table}[htbp]
\caption{Overall constituent-skill completion (TSCR, \%) on the complete 50-task Behavior-Skill benchmark. Task and Skill denote task and skill training, respectively.}
\label{tab:overall_tscr_50task}
\centering
\small
\setlength{\tabcolsep}{4pt}
\renewcommand{\arraystretch}{1.15}
\begin{tabular*}{\columnwidth}{@{\hspace{6pt}}>{\centering\arraybackslash}p{0.34\columnwidth} >{\centering\arraybackslash}p{0.18\columnwidth} >{\centering\arraybackslash}p{0.18\columnwidth} >{\centering\arraybackslash}p{0.12\columnwidth}@{\hspace{6pt}}}
\hline
\textbf{Model} & \textbf{Task} & \textbf{Skill} & \textbf{$\Delta$} \\ \hline
$\pi_{0.5}$ & 42.4 & 48.4 & +6.0 \\
GR00T N1.7 & 36.9 & 42.5 & +5.6 \\ \hline
\end{tabular*}
\vspace{-0.2cm}
\end{table}

Fig.~\ref{fig:over_skill_performance} reveals substantial variation in execution reliability across semantic skill types. Skills involving relatively simple spatial reasoning or object-state transitions, including Sweep Surface, Place Under, Hold, Spray, Release, and Move To, achieve consistently high success rates. Several manipulation skills, such as Place On, Place Next To, and Place In, attain moderate performance. In contrast, articulated-object interaction (Open Lid, Close Door, Close Lid), precise object manipulation (Pick Up From), and tool-use behaviors (Pour) remain among the lowest-performing skill types. Although the two evaluated policies differ in overall performance, they exhibit remarkably similar capability profiles, suggesting that these challenging skill types are shared bottlenecks across the two representative policies rather than unique to a particular architecture.

Skill training raises TSCR from 42.4\% to 48.4\% for $\pi_{0.5}$ and from 36.9\% to 42.5\% for GR00T N1.7 as shown in Table~\ref{tab:overall_tscr_50task}. However, these gains do not eliminate the low success rates observed for many semantic skill types. Even the best-performing setting achieves an average TSCR below 50\%, while several interaction skills remain rarely completed. These results indicate that constituent-skill capability remains substantially limited for the evaluated policies, despite the use of skill training.

\subsection{Comparison with Conventional Full-Task Evaluation}

\begin{figure*}[htbp]
\centering
\includegraphics[width=\textwidth]{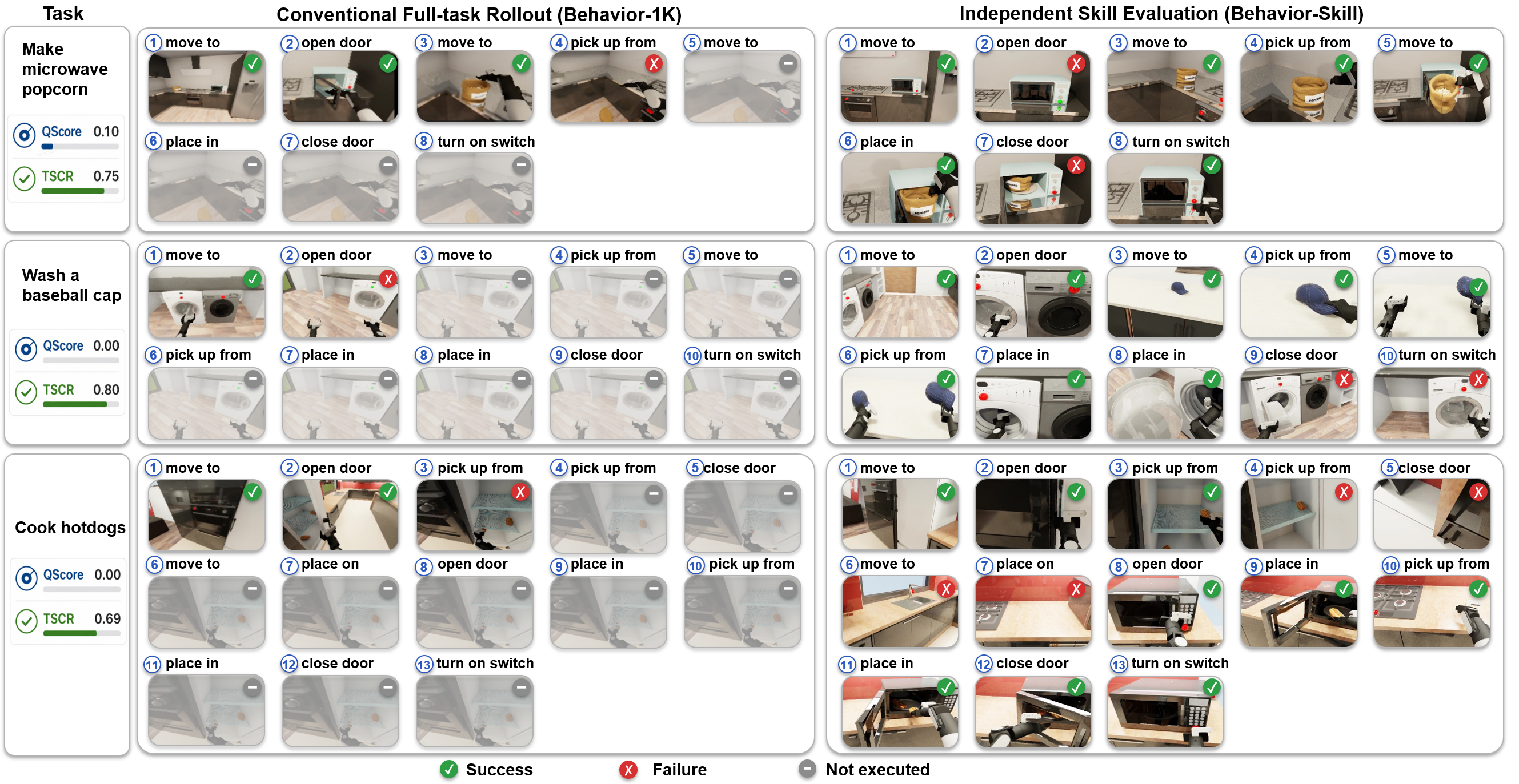}
\caption{Comparison of conventional full-task evaluation and Behavior-Skill on representative long-horizon tasks. Green, red, and gray markers indicate successful, failed, and unexecuted constituent skills, respectively.}
\label{fig:b1k_vs_bs}
\vspace{-0.2cm}
\end{figure*}

We compare Behavior-Skill with the conventional full-task evaluation on three representative long-horizon tasks using the $\pi_{0.5}-Task$ policy. The same policy is evaluated under the original BEHAVIOR-1K benchmark and the proposed Behavior-Skill benchmark. Since the original benchmark reports only task-level outcomes, the execution status of each constituent skill is manually identified from the recorded evaluation videos. 

Fig.~\ref{fig:b1k_vs_bs} reveals a clear difference between the two evaluation protocols. In the representative examples, intermediate failures prevent later constituent skills from being executed, leaving part of the task unevaluated. For example, in Make Microwave Popcorn, execution reaches the first five constituent skills, while the remaining three are never executed. Similar observations are found in Wash a Baseball Cap and Cook Hotdogs, where later constituent skills are not reached after intermediate failures. Consequently, only the executed portion of the task contributes to the reported task-level outcome, while later execution remains unobserved.

The comparison further highlights the difference between task-level and skill-level evaluation. Although Wash a Baseball Cap and Cook Hotdogs both obtain a QScore of 0\%, their TSCR values are 80.0\% and 69.0\%, respectively. Similarly, Make Microwave Popcorn achieves a QScore of 10.0\% with a TSCR of 75.0\%. These examples show that task-level outcomes and constituent-skill completion can differ substantially. In contrast, Behavior-Skill evaluates every constituent skill independently, allowing execution outcomes to be observed across the complete task.

\subsection{Task-Dependent Skill Capability}

\begin{table*}[htp]
\caption{Task Skill Completion Rate (TSCR, \%) on the 12-task subset under the Task and Skill settings. Values are reported as mean $\pm$ standard deviation over ten evaluation runs. $\Delta$ denotes the absolute improvement of Skill over Task in percentage points.}
\label{tab:12task_result}
\centering
\newcommand{\pmcell}[2]{\parbox[c]{\linewidth}{\centering #1\\$\pm$#2}}
\setlength{\tabcolsep}{3pt}
\renewcommand{\arraystretch}{1.8}
\begin{tabular*}{\textwidth}{
@{\extracolsep{\fill}}
p{1.20cm}
*{12}{>{\centering\arraybackslash}p{1.05cm}}
>{\centering\arraybackslash}p{0.80cm}
@{}}
\hline
\textbf{Setting}
& \shortstack[c]{Make\\Microwave\\Popcorn}
& \shortstack[c]{Hiding\\Easter\\Eggs}
& \shortstack[c]{Cook\\Hot\\Dogs}
& \shortstack[c]{Wash a\\Baseball\\Cap}
& \shortstack[c]{Picking\\Up\\Trash}
& \shortstack[c]{Turning\\On\\Radio}
& \shortstack[c]{Setting\\the\\Fire}
& \shortstack[c]{Bring\\Water}
& \shortstack[c]{Attach\\Camera\\to Tripod}
& \shortstack[c]{Hanging\\Pictures}
& \shortstack[c]{Putting\\Shoes\\on Rack}
& \shortstack[c]{Tidying\\Bedroom}
& \textbf{Mean} \\
\hline
Task
& \pmcell{50.4}{9.7}
& \pmcell{65.3}{8.2}
& \pmcell{49.8}{7.5}
& \pmcell{59.5}{7.9}
& \pmcell{57.8}{8.1}
& \pmcell{60.3}{12.1}
& \pmcell{63.3}{6.2}
& \pmcell{38.8}{11.1}
& \pmcell{58.0}{9.4}
& \pmcell{47.4}{10.1}
& \pmcell{36.3}{6.3}
& \pmcell{36.6}{7.6}
& 52.0 \\[4pt]

Skill
& \pmcell{70.4}{8.8}
& \pmcell{77.2}{7.1}
& \pmcell{62.8}{10.1}
& \pmcell{66.9}{6.7}
& \pmcell{71.6}{8.0}
& \pmcell{70.0}{11.3}
& \pmcell{68.6}{6.1}
& \pmcell{52.2}{11.0}
& \pmcell{57.3}{7.6}
& \pmcell{43.8}{13.0}
& \pmcell{55.4}{7.2}
& \pmcell{45.9}{9.5}
& 61.8 \\

$\Delta$ 
& +20.0
& +11.9
& +13.0
& +7.4
& +13.8
& +9.7
& +5.3
& +13.4
& -0.7
& -3.6
& +19.1
& +9.3
& +9.8 \\
\hline
\end{tabular*}
\vspace{-0.5cm}
\end{table*}

The benchmark-wide analysis in the previous section identifies the overall capability profile of constituent skills by aggregating execution outcomes across all tasks. This experiment further examines whether the execution reliability of the same semantic skill remains consistent under different task contexts. To this end, we fine-tune an additional $\pi_{0.5}$ model on a representative 12-task subset following the protocol of \cite{bai2025openpicomet}.  Each task is evaluated over ten independent rollouts, and all reported results are averaged across the ten runs to reduce rollout stochasticity.

\begin{figure}[!t]
\centering
\captionsetup[subfloat]{font=scriptsize}

\subfloat[Task]{%
    \includegraphics[width=0.98\columnwidth]
    {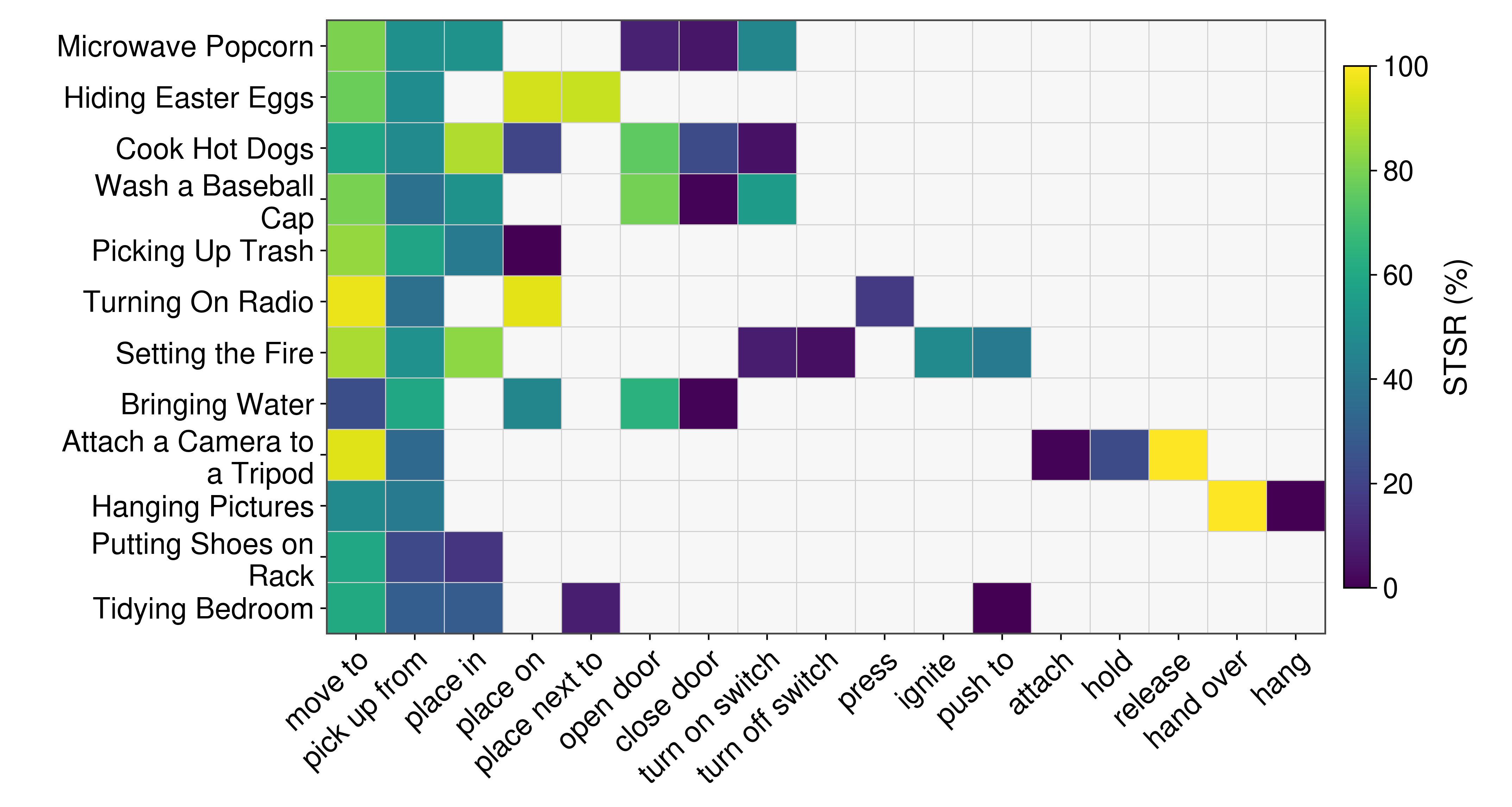}%
    \label{fig:stsr_task}%
}

\par\vspace{-0.5mm}

\subfloat[Skill]{%
    \includegraphics[width=0.98\columnwidth]
    {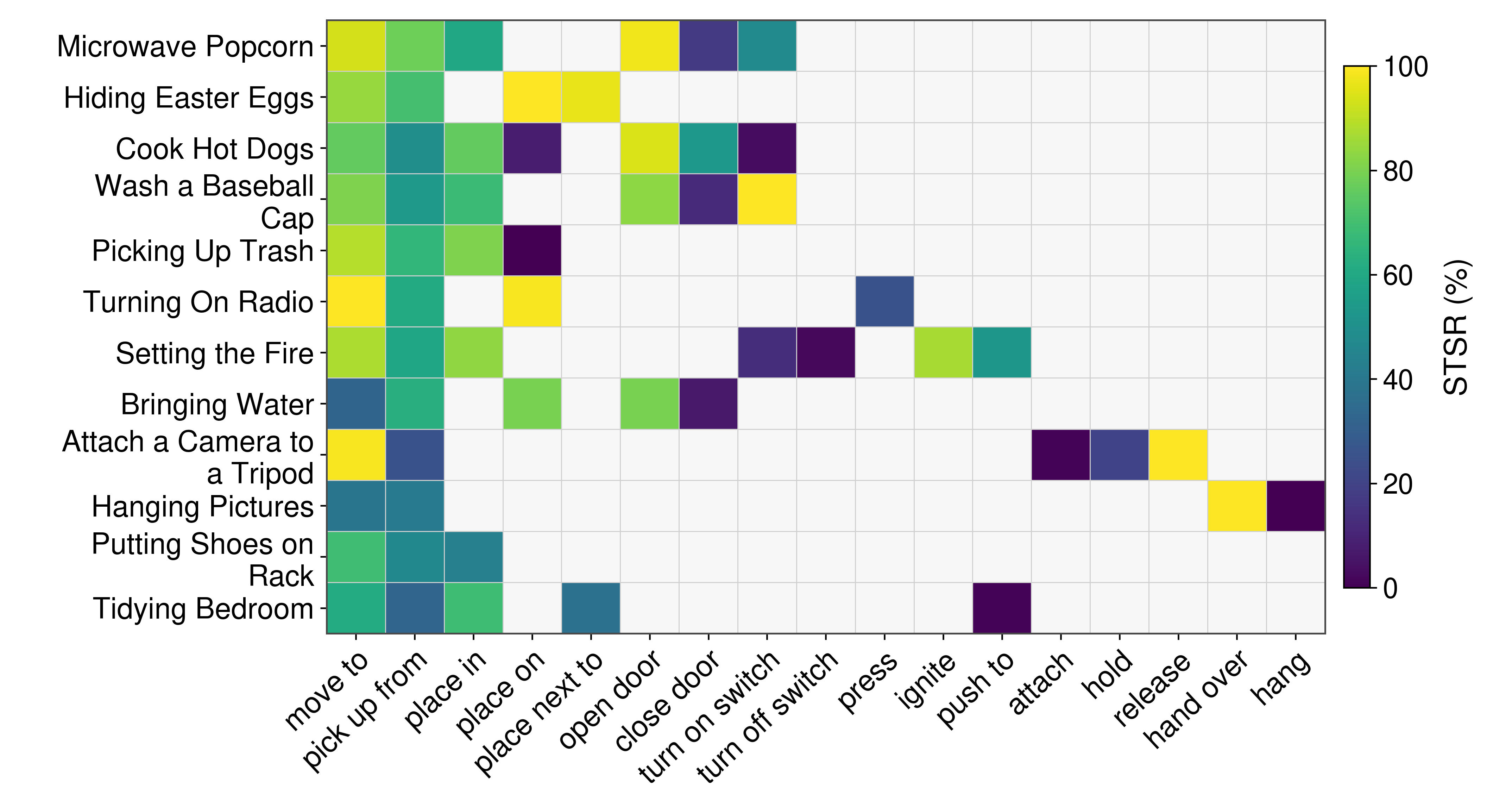}%
    \label{fig:stsr_skill}%
}

\caption{%
Task-wise Skill-Type Success Rate (STSR) on the representative
12-task subset under (a) the Task setting and (b) the Skill
setting. Color indicates STSR (\%), and blank cells denote skill
types absent from the corresponding task.%
}
\label{fig:stsr_heatmap}

\vspace{-5mm} 
\end{figure}

Task-wise skill completion differs substantially across the different activities. As shown in Table
~\ref{tab:12task_result}, TSCR ranges from 36.3\% to 65.3\% under the Task setting and from 43.8\% to 77.2\% under the Skill setting. Fig.~\ref{fig:stsr_heatmap} further decomposes these task-level differences into constituent-skill outcomes. Different tasks exhibit distinct distributions of semantic skill success rates.  For example, Make Microwave Popcorn consistently achieves high success rates for \textit{Move To} but low success rates for \textit{Open Door} and \textit{Close Door}, whereas Attach a Camera to a Tripod consistently achieves high success rates for \textit{Move To} and \textit{Release} but near-zero success for \textit{Attach}. Similar capability profiles are observed under both evaluation settings. 

The execution reliability of the same semantic skill also varies across different activities. As shown in \cref{fig:stsr_heatmap}, the success rate of the same semantic skill differs considerably across tasks. Under the Task setting (\cref{fig:stsr_task}), \textit{Open Door} ranges from 9.0\% in Make Microwave Popcorn to 78.0\% in Cook Hot Dogs and Wash a Baseball Cap, while \textit{Place In} ranges from 15.2\% in Putting Shoes on Rack to 88.0\% in Cook Hot Dogs. Comparable task-dependent variations are also observed under the Skill setting (\cref{fig:stsr_skill}). This suggests that a semantic skill label alone does not determine difficulty. Object geometry, target relation, and surrounding scene context also strongly affect execution reliability.

\section{Conclusion}

This paper presented Behavior-Skill, a fine-grained benchmark that establishes constituent skills as fundamental units for long-horizon task learning and evaluation. Built upon BEHAVIOR-1K, Behavior-Skill provides a skill dataset, intermediate-state restoration, and independent skill evaluation with skill success goals. Together with capability-oriented metrics, it enables detailed investigation of constituent-skill execution within complex long-horizon tasks. Experiments on representative VLA policies show that conventional task evaluation can hide intermediate skill failures. Meanwhile, skill execution remains limited across the evaluated policies. These findings suggest that reliable execution of skills remains a key challenge for long-horizon mobile manipulation tasks.

Behavior-Skill provides a complementary perspective to conventional full-task evaluation. Specifically, it focuses on independent skill evaluation and does not cover task planning and automatic task decomposition, or the sequential dependencies between constituent skills during execution. Future work will extend Behavior-Skill toward planning-aware and sequential skill evaluation, supporting the development of more capable long-horizon embodied systems.

\bibliographystyle{IEEEtran}
\bibliography{bare_jrnl_new_sample5}

\newpage

\vfill

\end{document}